\documentclass[11pt,a4paper]{article}
\usepackage{authblk}
\usepackage[hyperref]{emnlp2018}
\usepackage{times}
\usepackage{latexsym}

\usepackage{url}
\usepackage[utf8]{inputenc}
\usepackage{amsmath}
\usepackage{txfonts}
\usepackage[disable]{todonotes}
\usepackage{textcomp}

\usepackage{hyphenat}
\aclfinalcopy 
\def\aclpaperid{1117} 

\title{Joint Aspect and Polarity Classification for Aspect-based Sentiment Analysis with End-to-End Neural Networks}

\author[1]{Martin Schmitt}
\author[1,2]{Simon Steinheber}
\author[2]{Konrad Schreiber}
\author[1]{Benjamin Roth}

\affil[1]{Center for Information and Language Processing\\ LMU Munich, Germany}
\affil[2]{MaibornWolff GmbH, Munich, Germany}
\affil[ ]{\texttt{martin@cis.lmu.de}}

\date{}

\newcommand{\cset}[1]{\left\{ \, #1 \, \right\}}

\DeclareMathOperator*{\argmax}{arg\,max}

\newcommand{\textapprox}{\raisebox{0.5ex}{\texttildelow}}

\begin{document}

\maketitle

\begin{abstract}
  In this work, we propose a new model for aspect-based sentiment analysis.
  In contrast to previous approaches, we jointly model the detection of aspects and the classification of their polarity in an end-to-end trainable neural network.
  We conduct experiments with different neural architectures and word representations on the recent GermEval 2017 dataset.
  We were able to show considerable performance gains by using the joint modeling approach in all settings compared to pipeline approaches.
  The combination of a convolutional neural network and fasttext embeddings outperformed the best submission of the shared task in 2017, establishing a new state of the art.
\end{abstract}


\section{Introduction}
Sentiment analysis \cite{pang08} is the automatic detection of the sentiment expressed in a piece of text.
Typically, this is modeled as a classification task with at least two classes (positive, negative), sometimes extended to three (neutral) or more fine-grained categories.
Aspect-based sentiment analysis (ABSA) aims at a finer analysis, i.e.\ it requires that certain aspects of an entity in question be distinguished and the sentiment be classified with regard to each of them.
An example can be seen in \autoref{fig:absa_example}.

This introduces several new challenges.
First, labeled data, which are needed to train statistical models, are more difficult to obtain.
Therefore the amount of available training data is limited.
Thus a good model for ABSA has to make the best possible use of the available data.
Second, the detection of the subset of aspects that occur in a given piece of text is non-trivial.
Errors introduced at this stage severely limit the performance on the overall ABSA task.
Third, the general sentiment and the sentiment of each aspect can each be completely different from each other (cf.\ \autoref{fig:absa_example}).
This means that a model has to be able to distinguish aspects in the text and make independent decisions for each of them.

We want to address each of these challenges by
(1) leveraging unlabeled data by modeling word representations and
(2) modeling aspect detection and classification of their polarity jointly in an end-to-end trainable system.

We evaluate our approach on the GermEval 2017 data, i.e.\ customer reviews about \emph{Deutsche Bahn AG} on social media. We particularly address subtask C as the typical setting where two pieces of information have to be detected from raw text:
\begin{enumerate}
\item Which aspects are mentioned?
\item For each mentioned aspect, what is the polarity of its sentiment?
\end{enumerate}

From the new state-of-the-art results we obtain, we conclude that
modeling of word representations and joint modeling of aspects and polarity
have not yet received the attention they deserve.

\begin{figure}
  \underline{German:}
  Alle so ``Yeah, Streik beendet'' Bahn so ``Okay, dafür werden dann natürlich die Tickets teurer'' Alle so ``Können wir wieder Streik haben?''
  
  \vspace{0.2cm}

  \underline{Translation:}
  Everybody's like ``Yeah, strike's over'' Bahn goes ``Okay, but therefore we're going to raise the prices'' Everybody's like ``Can we have the strike back?''

  \vspace{1em}
  \begin{tabular}{ll}
    General sentiment: & neutral\\
    Aspect sentiment: & Ticket purchase:negative\\
                       & General:positive\\[.7em]
  \end{tabular}

  \caption{Example sentence with contained aspects and their polarity.}
  \label{fig:absa_example}
\end{figure}



\section{Related Work}

Two recent shared tasks address ABSA:
SemEval 2016 Task 5 \cite{semeval16} and GermEval 2017 \cite{germeval17}.
The SemEval dataset is extremely small.
The English laptop reviews, e.g., only contain 395 training instances for the prediction of 88 aspect categories and their polarities.
Because of this sparsity, top-ranked systems rely on feature engineering and hand-crafted rules.
GermEval is a larger dataset (\textapprox 20K training instances, 20 aspect categories) and thus suited for our goal of evaluating the quality of fully automatic methods for learning aspect and polarity predictions.
Furthermore the top systems at SemEval 2016, XRCE \cite{xrce_semeval16} and IIT-TUDA \cite{iit-tuda_semeval16}, not only rely heavily on feature engineering but also separate the tasks of aspect detection and aspect polarity classification into two different parts of their pipeline.

The winners of GermEval 2017 rely on neural methods \cite{lee_germeval17}.
They try to link all aspects to a sequence of tokens and model the task as a sequence labeling problem.
This leads to problems because some aspects are not assigned to any token but still have to be detected and classified.
Our approach always considers the complete document and produces the set of all detected aspects at once.
Although \citet{lee_germeval17} incorporate some aspects of multi-task learning, the prediction of aspect category and polarity remains separated in each of their approaches.
In our work, we show that a joint learning of these two tasks achieves better performance.
The approach by \citet{lee_germeval17} also relies more heavily on external sources than ours.
While we only collected a corpus of \textapprox 113K unlabeled German tweets, \citet{lee_germeval17} include annotated English data as well as a much larger unlabeled German corpus (Wikipedia, cf.\ \citet{al-rfou13}) in their setting.

\citet{ruder16} propose another neural model for ABSA.
Similarly to the approaches mentioned above, they assume that aspects have already been detected by some other system in a pipeline architecture, and they concentrate on polarity classification on the sentence level by unifying information from other sentences on the document level.
We compare ourselves to this baseline and show improvements over pipeline approaches.


\section{Proposed Model}
\label{sec:models}

\subsection{Embedding Algorithm}
Word2vec skip-gram \cite{word2vec} is a widely used algorithm to obtain pretrained vector representations for input words.
Notably, \citet{lee_germeval17} use it for their experiments on the GermEval data.
FastText \cite{fasttext} works in a similar fashion but has the advantage of incorporating subword information in the embedding learning process.
So it can not only learn similar embeddings for word forms sharing a common stem
but also generate embeddings for unseen words in the test set by combining the learned character ngram embeddings.
This can be crucial when dealing with a morphologically rich language such as German.
Glove \cite{glove} --- similar to word2vec --- does not incorporate character-level information, but uses global rather than local information to learn its word embeddings.

We have trained each of these embedding learning algorithms on a corpus of \textapprox 113K tweets mentioning at least one of \texttt{@DB\_info} and \texttt{@DB\_Bahn}, two official accounts of \emph{Deutsche Bahn AG} offering information and replying to questions. We collected these tweets specifically to build a document collection that is closely related to the domain of GermEval 2017.
We also included the GermEval training set for the embedding training.

\subsection{Pipeline LSTM (baseline)}
\label{sec:pipeline-lstm}

We compare our proposed approach to the model described in \cite{ruder16}.
They first encode each sentence with glove word embeddings and a bidirectional LSTM \cite{hochreiter97}.
Then this output is concatenated with an embedding of the aspect addressed in the current sentence and finally fed in a document-level BiLSTM.
As we are dealing with social media texts, our documents are already very short.
So we do not split them into shorter units (sentences).
Therefore the second hierarchy level of \cite{ruder16},
that combines the output of consecutive sentences in a document,
is superfluous and omitted in our experiments.
In all other aspects --- including hyperparameters --- we do as \cite{ruder16}, i.e.\ we duplicate a tweet for each aspect detected in it, concatenate an aspect embedding of size $15$ to the output of the BiLSTM encoder, use dropout of $0.5$ after the embedding layer and after LSTM cells, and apply a gradient clipping norm of 5.
As \citet{ruder16} rely on the detected aspects to be given at test time, for a realistic comparison, we feed in the aspects as detected by the strong GermEval baseline system based on support vector machines.
The so obtained system serves as our first baseline, representing a state-of-the-art pipeline system.

\subsection{End-to-end LSTM}
We modify the pipeline model described in the last section as follows:
the aspect detection is integrated into the neural network architecture permitting an end-to-end optimization of the whole model during training.
This is achieved by formatting the classifier output as a vector $\mathbf{z} \in \cset{0, 1, 2, 3}^{|A|}$, where $A$ is the set of all 20 aspects (e.g., \emph{General, Ticket purchase, Design, Safety, $\dots$}). 
This corresponds to predicting one of the four classes \emph{N/A}, \emph{positive}, \emph{negative} and \emph{neutral} for each aspect.
Specifically, we obtain a hidden representation of an input document $X$ in the following manner:
\begin{equation}
  \label{eq:1}
  \mathbf{v} = \textit{DO}(\textit{BiLSTM}(\textit{DO}(\textit{embed}(X))))
\end{equation}
where $\textit{embed} \in \cset{\textit{word2vec}, \textit{glove}, \textit{fasttext}}$\\
and DO = dropout \cite{hinton12}.

\noindent The design choices for the BiLSTM in this step remain the same as in the baseline model.

Then, we transform the feature vector $\mathbf{v}$ extracted from the text $X$ to a score vector $\hat{y}^{(a)}$ for each aspect $a\in A$ and apply softmax normalization:
\begin{equation}
  \label{eq:2}
  \hat{y}^{(a)} = \textit{softmax}(W^{(a)} \mathbf{v} + b^{(a)})
\end{equation}
where
\begin{equation}
  \label{eq:7}
  \textit{softmax}(\mathbf{x})_i = \frac{\textit{exp}(x_i)}{\sum_{k=0}^3 ~ \textit{exp}(x_k)} ~~~ \text{for}~i = 0, \dots, 3
\end{equation}
Thus for each aspect, we predict its presence or absence as well as its polarity in one step:
\begin{equation}
  \label{eq:6}
  z^{(a)} = \argmax_i ~ \hat{y}_i^{(a)}
\end{equation}
The loss is simply the cross entropy summed over all aspects:
\begin{equation}
  \label{eq:3}
  {L}(\theta) = \sum_{a\in A} ~ H(y^{(a)},~ \hat{y}^{(a)})
\end{equation}
with
\begin{equation}
  \label{eq:4}
  H(y,~ \hat{y}) = -\sum_i ~ y_i \cdot \textit{log}(\hat{y}_i)  
\end{equation}

\subsection{End-to-End CNN}

\begin{figure}
  \centering
  \includegraphics[width=\columnwidth]{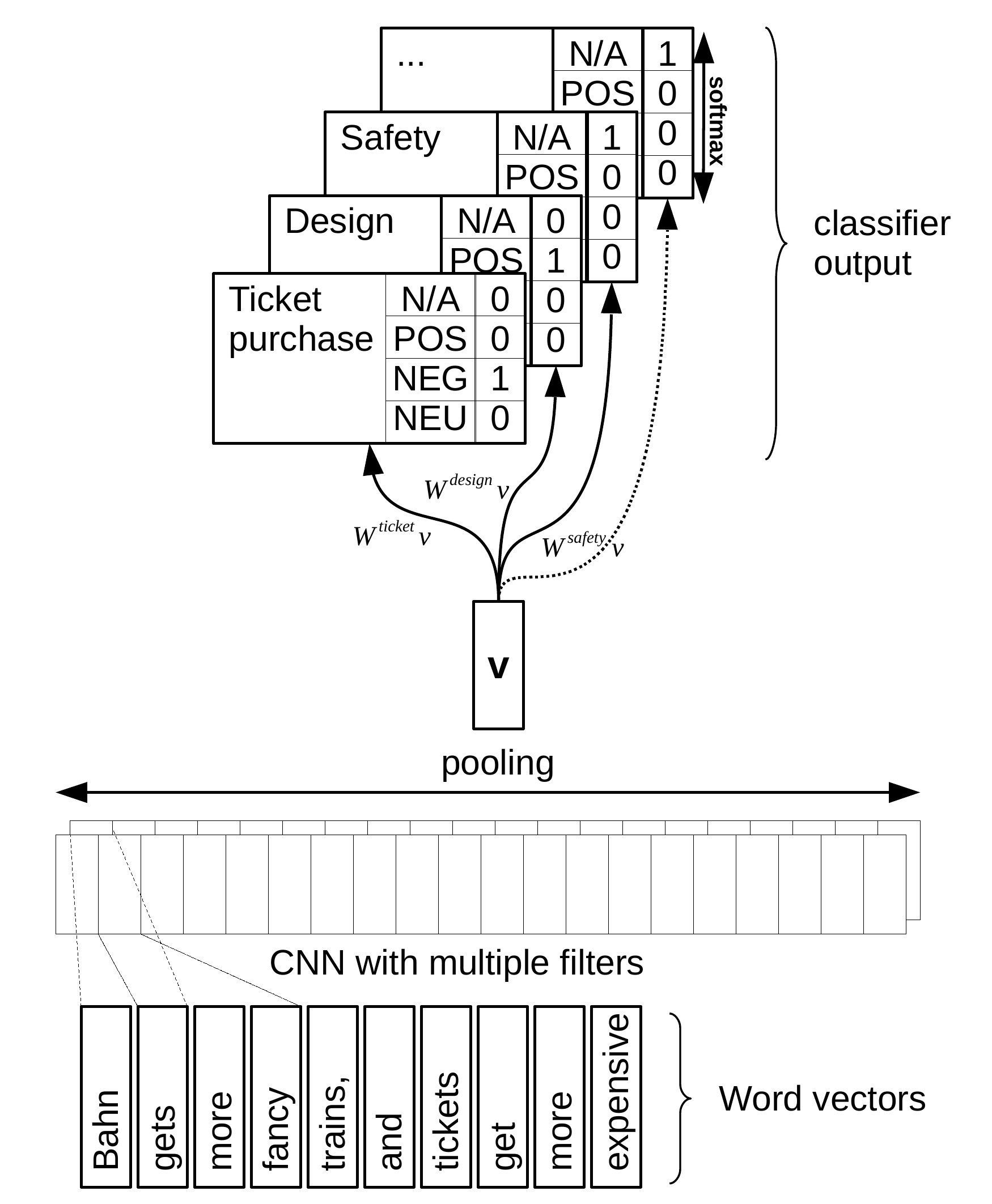}
  \caption{Schematic view of end-to-end CNN architecture.}
\end{figure}

Keeping the formalization as an end-to-end task, we replace the BiLSTM by a convolutional neural network (CNN) as described in \cite{kim14}. As in their setting \emph{CNN-non-static}, we use $300$-dimensional word embeddings, a max-over-time pooling operation, filter sizes of $3, 4, 5$, and dropout with a rate of $0.5$ (as before). We use ReLu ($f(x) = \textit{max}(0, x)$) as our activation function, and $300$ filters of each size, a number also found in related work on sentiment analysis \cite{dosSantos14}.
Following \cite{kim14}, we do not apply dropout after the embedding layer:
\begin{equation}
  \label{eq:5}
  \mathbf{v} = \textit{DO}(\textit{CNN}(\textit{embed}(X)))
\end{equation}
With \autoref{eq:5} replacing \autoref{eq:1}, the aspectwise classification for the end-to-end CNN then follows the same definitions as described in the previous section.
\todo{add graphic for CNN architecture}

\subsection{Pipeline CNN}

In order to compare the effects of joint end-to-end and pipeline approaches across neural architectures,
we also include an experiment where the CNN model from the previous section
replaces the BiLSTM in the pipeline setting described in section \ref{sec:pipeline-lstm}.


\section{Experiments}
We conduct our experiments on the GermEval 2017 data \cite{germeval17}, i.e.\ customer feedback about \emph{Deutsche Bahn AG} on social media, microblogs, news, and Q\&{}A sites.
The data were collected over the time of one year and manually annotated, resulting in a main dataset of about 26K documents, divided into a training, development, and test set using a random 80\%{}/10\%{}/10\%{} split.
About 1,800 documents from the last 3 months of the data collection period constitute a diachronic test set that can be used to test the robustness of a system over time.
We keep the proposed data split and filter out training instances where the same aspect category was assigned two different polarity classes (which affects approximately $4\%{}$ of the data).
The development and test data remain the same.

\begin{table*}[htb]
  \centering
  \begin{tabular}{ll|c|c|c}
    &&development set&synchronic test set & diachronic test set\\\hline
    Pipeline LSTM & + word2vec & .350 & .297 & .342\\
    End-to-end LSTM & + word2vec & .378 & .315 & .383\\
    Pipeline CNN & + word2vec & .350 & .298 & .343 \\
    End-to-end CNN & + word2vec & .400 & .319 & .388\\\hline
    Pipeline LSTM & + glove & .350 & .297 & .342\\
    End-to-end LSTM & + glove & .378 & .315 & .384\\
    Pipeline CNN & + glove & .350 & .298 & .342 \\
    End-to-end CNN & + glove & .415 & .315 & .390\\\hline
    Pipeline LSTM & + fasttext & .350 & .297 & .342\\
    End-to-end LSTM & + fasttext & .378 & .315 & .384\\
    Pipeline CNN & + fasttext & .342 & .295 & .342 \\
    End-to-end CNN & + fasttext & \textbf{.511} & \textbf{.423} & \textbf{.465} \\
    \hline
    \multicolumn{2}{l|}{majority class baseline} & -- & .315 & .384\\
    \multicolumn{2}{l|}{GermEval baseline} & -- & .322 & .389\\
    \multicolumn{2}{l|}{GermEval best submission} & -- & .354 & .401\\
  \end{tabular}
  \caption{Results on the GermEval data, \textbf{aspect + sentiment} task. Micro-averaged F1-scores for both aspect category and aspect polarity classification as computed by the GermEval evaluation script. In the bottom part of the table, we report results from \cite{germeval17}.}
  \label{tab:results}
\end{table*}


\begin{table*}[htb]
  \centering
  \begin{tabular}{ll|c|c|c}
    &&development set&synchronic test set&diachronic test set\\\hline
    End-to-end LSTM & + word2vec & .517 & .442 & .455 \\
    End-to-end CNN & + word2vec & .521 & .436 & .470 \\\hline
    End-to-end LSTM & + glove & .517 & .442 & .456 \\
    End-to-end CNN & + glove & .537 & .457 & .480 \\\hline
    End-to-end LSTM & + fasttext & .517 & .442 & .456 \\
    End-to-end CNN & + fasttext & \textbf{.623} & \textbf{.523} & \textbf{.557} \\
    \hline
    \multicolumn{2}{l|}{majority class baseline} & -- & .442 & .456\\
    \multicolumn{2}{l|}{GermEval baseline} & -- & .481 & .495 \\
    \multicolumn{2}{l|}{GermEval best submission} & -- & .482 & .460\\
  \end{tabular}
  \caption{Micro-averaged F1-scores for the prediction of \textbf{aspect categories only} (i.e.\ without taking polarity into account at all) as computed by the GermEval evaluation script. The results in the bottom part of the table are taken from \cite{germeval17}.}
  \label{tab:aspect}
\end{table*}


We choose our hyperparameters based on the development data using the following procedure:
we train initial models with a hyperparameter setting based on values we found in the literature, 
stochastic gradient descent with a learning rate of $0.01$ (as in \citet{dosSantos14}) and a mini-batch size of $10$ (as in \citet{ruder16}).
For the best-performing CNN and LSTM architectures (end-to-end + fasttext), we then refine the learning rate and batch size on the development data using random search in the range $\cset{0.001, 0.003, 0.01, 0.03, 0.1}$ for learning rate and $\cset{5, 10, 20}$ for batch size.
For the CNN setting, this results in a learning rate of $0.03$ and a batch size of $5$ (which we then use for all CNN architectures in the final experiments).
For the LSTM setting, this results in a learning rate of $0.01$ and a batch size of $10$ (which we then use for all LSTM architectures).

Training our models takes between 1-3 minutes per epoch on a GeForce GTX 1080 GPU,
the end-to-end CNN being the fastest model to train.


\section{Discussion}

\paragraph{Aspect polarity}

\autoref{tab:results} shows the results of our experiments, as well as the results of our strong baselines.
Note that the majority class baseline already provides good results.
This is due to highly unbalanced data;
the aspect category ``\emph{Allgemein}'' (\emph{``general''}), e.g., constitutes 61.5\%{} of the cases.
This imbalance makes the task even more challenging.

Over all architectures, we observe a comparable or better performance when using fasttext embeddings instead of word2vec or glove.
This backs our hypothesis that subword features are important for processing the morphologically rich German language.
Leaving everything else unchanged, we can furthermore see an increase in performance for all settings, when switching from the pipeline to an end-to-end approach.
The best performance (marked in bold) is achieved by a combination of CNN and FastText embeddings, which outperforms the highly adapted winning system of the shared task.

\paragraph{Aspect category only}

Even though our architectures are designed for the task of joint prediction of aspect category and polarity,
we can also evaluate them on the detection of aspect categories only.
\autoref{tab:aspect} shows the results for this task.
First of all, we can see that the SVM-based GermEval baseline model has very decent performance
as it is practically on par with the best submission for the synchronic
and even outperforms the best submission on the diachronic test set.
It is therefore well-suited to serve as input to the pipeline LSTM model we compare with in our main task.

Comparing our architectures, we see again that fasttext embeddings always lead to equal or better performance.
And even though we do not directly optimize our models for this task only,
our best model (CNN+fasttext) outperforms all baselines, as well as the GermEval winning system.

\paragraph{Impact of domain-specific corpus}

We compare the domain-specific FastText embeddings to FastText embeddings trained on Wikipedia\footnote{Downloaded from \url{https://github.com/facebookresearch/fastText/blob/master/pretrained-vectors.md}.},
which is approximately 100 times the size of our domain-specific corpus.
We report the results in \autoref{tab:wiki}. The embeddings trained on Wikipedia show slightly lower performance on the dev set but slightly higher or equal performance on the test sets.
We conclude that the main positive impact of FastText stems from its capability to model subword information and that a large domain-independent corpus or a small domain-specific corpus lead to similar performance gains.

\begin{table}
  \centering
  \begin{tabular}{l|c|c|c}
    & dev & synchr.\ test & diachr.\ test\\
    \hline
    aspect + sent. & .502 & .423 & .465\\
    aspect only & .610 & .544 & .571
  \end{tabular}
  \caption{Results of the end-to-end CNN model with fasttext embeddings trained on the German Wikipedia.}
  \label{tab:wiki}
\end{table}


\section{Conclusion}
We have presented a new approach to ABSA.
By solving the two classification problems (aspect categories + aspect polarity) inherent to ABSA in a joint manner,
we observe significant performance gains for both of these tasks on the GermEval 2017 data.
Our experiments also showed that word representations leveraging subword information
are crucial for a challenging task like ABSA in a morphologically rich language,
such as German.
Furthermore we observed consistently better performance of CNN architectures in otherwise comparable scenarios,
which suggests that CNNs cope better with the irregularities of user-written texts on social media,
a research question we leave to future work.
By establishing a new state of the art in aspect detection and polarity classification,
we provide a new practical baseline for future research in this area.


\section*{Acknowledgments}

We would like to thank the anonymous reviewers for their valuable input.
This work was partially supported by the European Research Council,
Advanced Grant \# 740516 NonSequeToR, and by a Ph.D. scholarship awarded to the first author by the German Academic Scholarship Foundation (Studienstiftung des deutschen Volkes).


\bibliographystyle{acl_natbib_nourl}
\bibliography{references}

\end{document}